\documentclass[twoside]{article}
\usepackage[accepted]{aistats2012}

\usepackage{graphicx} 
\usepackage{algorithm}
\usepackage{algorithmic}
\usepackage{epsf}
\usepackage{amsmath}
\usepackage{amssymb}
\usepackage{graphicx}
\usepackage{tabularx}
\usepackage{rotating}
\usepackage{url}
\usepackage{mathrsfs}
\usepackage{Definitions}
\def\tps{^{\mathsf{T}}}

\begin{document}

%

%

\twocolumn[

\aistatstitle{Two-Manifold Problems}

\aistatsauthor{ Byron Boots \And Geoffrey J. Gordon }

\aistatsaddress{Machine Learning Department\\ Carnegie Mellon University\\ Pittsburgh, PA 15213\\ beb@cs.cmu.edu \And Machine Learning Department\\ Carnegie Mellon University\\ Pittsburgh, PA 15213\\ ggordon@cs.cmu.edu} ]

\begin{abstract}
  Recently, there has been much interest in spectral approaches to
  learning manifolds---so-called kernel eigenmap methods.  These
  methods have had some successes, but their applicability is limited
  because they are not robust to noise.  To address this limitation,
  we look at \emph{two-manifold problems}, in which we simultaneously
  reconstruct two related manifolds, each representing a different
  view of the same data.  By solving these interconnected learning
  problems together and allowing information to flow between them,
  two-manifold algorithms are able to succeed where a non-integrated
  approach would fail: each view allows us to suppress noise in the
  other, reducing bias in the same way that an \emph{instrumental
    variable} allows us to remove bias in a {linear}
  dimensionality reduction problem.  We propose a class of algorithms
  for two-manifold problems, based on spectral decomposition of
  cross-covariance operators in Hilbert space.  Finally, we discuss situations
  where two-manifold problems are useful, and demonstrate that solving
  a two-manifold problem can aid in learning a nonlinear dynamical
  system from limited data.
\end{abstract}

\section{Introduction}
Manifold learning algorithms are non-linear methods for embedding a
set of data points into a low-dimensional space while preserving local
geometry. Recently, there has been a great deal of interest in
spectral approaches to learning manifolds. These \emph{kernel
  eigenmap} methods include Isomap~\cite{Tenenbaum00}, Locally Linear
Embedding (LLE)~\cite{lleRoweis}, Laplacian Eigenmaps
(LE)~\cite{Belkin02}, Maximum Variance Unfolding
(MVU)~\cite{Weinberger04}, and Maximum Entropy Unfolding
(MEU)~\cite{lawrence-aistats11}.  These approaches can be viewed as
kernel principal component analysis~\cite{kernelPCA} with specific
choices of manifold kernels~\cite{Ham03}: they seek a small set of
latent variables that, through a nonlinear mapping, explains the
observed high-dimensional data.

Despite the popularity of kernel eigenmap methods, 
they are limited in one important respect: they generally
only perform well when there is \emph{little or no noise}. Several
authors have attacked this problem using methods including
neighborhood smoothing~\cite{ChenYL08} and robust principal components
analysis~\cite{ZhanY09,ZhanY11}, with some success under limited noise.
Unfortunately, the problem is fundamentally ill posed without some
sort of side information about the true underlying signal: by design,
manifold methods will recover extra latent dimensions which
``explain'' the noise.

We take a different approach to the problem of learning manifolds from
noisy observations. We assume access to a set of \emph{instrumental
  variables}: variables that are correlated with the true latent
variables, but uncorrelated with the noise in observations. Such
instrumental variables can be used to separate signal from noise, as
described in Section~\ref{sec:crosscov}. 
Instrumental variables have been used to allow consistent estimation
of model parameters in many statistical learning problems,
including linear regression~\cite{Pearl2000}, principal component
analysis~\cite{Jolliffe2002}, and temporal difference
learning~\cite{Bradtke96linearleast-squares}. Here we extend the scope
of this technique to manifold learning.  We will pay particular
attention to the case of observations from \emph{two} manifolds, each of
which can serve as instruments for the other. We call such problems {\bf two-manifold problems}.

In one-manifold problems, when the noise variance is significant
compared to the manifold's geometry, kernel eigenmap methods are
biased: they will fail to recover the true manifold, even in the limit
of infinite data.  Two-manifold methods, by contrast, can succeed in
this case: we propose algorithms based on 
spectral decompositions related to \emph{cross-covariance operators}
in a tensor product space of two 
different manifold kernels, and show that the instrumental variable
idea suppresses noise in practice.  We also examine some theoretical
properties of two-manifold methods: we argue consistency for a special
case in
Sec.~\ref{sec:KSVD}, but we leave a full theoretical analysis of these
algorithms to future work.



The benefits of the spectral approach to two-manifold problems are
significant: first, the 
spectral decomposition naturally solves the manifold alignment problem
by finding low-dimensional manifolds (defined by the left and right
eigenvectors) that explain both sets of observations. Second, the
connection between manifolds and covariance operators opens
the door to solving more-sophisticated machine learning problems with
manifolds, including nonlinear analogs of canonical correlations
analysis (CCA) and reduced-rank regression (RRR).

As an example of this last point, \emph{subspace
  identification} approaches to learning
non-linear dynamical systems depend critically on instrumental
variables and the spectral decomposition of (potentially infinite-dimensional)
covariance
operators~\cite{zhang09,Siddiqi10a,Boots2010b,Song:2010fk,Boots2011a}.
Two-manifold problems are a natural fit: by
relating the spectral decomposition to our two-manifold method,
subspace identification techniques can be forced to identify a \emph{manifold}
state space, and consequently, to learn a dynamical system that is both accurate and interpretable, 
outperforming  the current state of the art.

\section{Preliminaries}
\label{sec:background}

We begin by looking at two well-known classes of nonlinear dimensionality reduction: kernel principal component analysis (kernel PCA) and manifold learning.

\subsection{Kernel PCA}
\label{sec:KPCA}
Kernel PCA~\cite{kernelPCA} is a generalization of principal component analysis~\cite{Jolliffe2002}: we first map our $d$-dimensional inputs $x_1,\hdots,x_n \in \mathbb{R}^d$
to a higher-dimensional feature space $\mathcal F$ using a feature
mapping ${\bf \phi}: \mathbb{R}^d \to \mathcal{F}$, and {then}
find the principal components in this new space.  If the features are
sufficiently expressive, kernel PCA can find structure that
regular PCA misses.  However, if $\mathcal{F}$ is high- or
infinite-dimensional, the straightforward approach to PCA, via an eigendecomposition of a covariance matrix, is in general intractable. Kernel
PCA overcomes this problem by assuming that $\mathcal F$ is a
reproducing-kernel Hilbert space (RKHS), and that the feature mapping
$\phi$ is 
\emph{implicitly} defined via an efficiently-computable kernel
function $K({\bf x,x'}) = \langle{\bf \phi}({\bf x}), {\bf \phi}({\bf
  x'})\rangle_\mathcal{F}$. Popular kernels include the linear kernel
$K({\bf x,x'}) = {\bf x} \cdot {\bf x'}$ (which identifies the feature
space with the input space) and the RBF kernel $K({\bf x,x'}) =
\exp(-\gamma\|{\bf x}-{\bf
  x'}\|^2/2)$.

Conceptually, if we define an infinitely-tall ``matrix'' with columns
$\phi({\bf x}_i)$, ${\bf \Phi} = (\phi({\bf x}_1), \ldots, \phi({\bf
    x}_n))$, our goal is to recover the eigenvalues and eigenvectors
of the centered \emph{covariance operator} $\hat{\bf \Sigma}_{XX} =
\frac{1}{n}{\bf \Phi H\Phi\tps}$, where $\bf H$ is the centering
matrix ${\bf H} = {\bf I}_{n} - \frac{1}{n}{\bf 11}\tps$.  To avoid
working with the high- or infinite-dimensional covariance operator, we
instead define the \emph{Gram matrix} ${\bf G} =\frac{1}{n}{\bf \Phi\tps\Phi}$.  The
nonzero eigenvalues of the centered covariance $\hat{\bf \Sigma}_{XX}$ are
the same as those of the centered Gram matrix ${\bf HGH}$.  And, the
corresponding unit-length eigenvectors of $\hat{\bf \Sigma}_{XX}$ are given by ${\bf
  \Phi H}{\bf v}_i\lambda^{-1/2}_i$, where $\lambda_i$ and ${\bf
  v}_i$ are the eigenvalues and eigenvectors of ${\bf
HGH}$~\cite{kernelPCA}.
%
%
If data weights $\bf P$ (a diagonal matrix) are present, we can instead decompose
the \emph{weighted} centered Gram matrix $\bf PHGHP$. (Note that, perhaps confusingly, we center first and multiply by the weights only after centering; the reason for this order will become clear below, in Section~\ref{sec:Manifold}.)

\subsection{Manifold Learning}
\label{sec:Manifold}
Spectral algorithms for manifold learning, sometimes called 
kernel
eigenmap methods, include Isomap~\cite{Tenenbaum00}, Locally Linear
Embedding (LLE)~\cite{lleRoweis}, Laplacian Eigenmaps (LE)~\cite{Belkin02}, and Maximum
Variance Unfolding (MVU)~\cite{Weinberger04}. 
These methods seek a nonlinear function that maps a high-dimensional
set of data points to a lower-dimensional space while preserving the
manifold on which the data lies. 
The main
insight behind these methods is that large distances in input space are often meaningless
due to the large-scale curvature of the manifold; so, ignoring these
distances can lead to a significant improvement in dimensionality
reduction by ``unfolding'' the manifold.

Interestingly, these algorithms can be viewed as special cases of kernel PCA where the Gram matrix $\bf G$ is constructed over the finite domain of the training data in a particular way~\cite{Ham03}.
Specifically, kernel eigenmap methods first induce a neighborhood
structure on the data, capturing a notion of local geometry via a
graph, where nodes are data points and edges are neighborhood
relations. Then they solve an eigenvalue or related problem based on
the graph to embed the data into a lower dimensional space while
preserving the local relationships. Here we just
describe LE; the other methods have a similar intuition, although the
details and performance characteristics differ.


%


In LE, neighborhoods are summarized by an adjacency
matrix $\bf W$, computed by nearest neighbors: element $w_{i,j}$ is
nonzero whenever the $i$th point is one of the nearest
neighbors of the $j$th point, or vice versa. Non-zero weights are
typically either set to 1 or computed according to a Gaussian RBF
kernel: $w_{i,j} = \exp(-\gamma\|{\bf x}_i - {\bf x}_j\|^2/2)$.  
Now let ${\bf
  S}_{i,i} = \sum_{j} w_{i,j}$, and set ${\bf C} = {\bf S}^{-1/2}{\bf
  (W-S)}{\bf S}^{-1/2}$.
Finally, eigendecompose $\bf C$ to get a low
dimensional embedding of the data points, discarding the top
eigenvector (which is trivial): if $\bf C = V\Lambda V\tps$, the embedding is
${\bf S}^{1/2}{\bf V}_{2:k+1}$.
To relate LE to kernel PCA,
note that
$\bf G = W-S$ is already centered, i.e., $\bf G = HGH$.  So, we can view LE
as performing weighted kernel PCA, with Gram matrix $\bf G$ and data
weights ${\bf P=S}^{-1/2}$.%
\footnote{The original algorithm actually
looks for the \emph{smallest} eigenvalues of $\bf -C$, which is
equivalent.  The matrix $\bf -G = \bf S-W$ is the \emph{graph Laplacian} for
our neighborhood graph, and $\bf-C$ is the \emph{weighted} graph
Laplacian---hence the algorithm name ``Laplacian eigenmaps.''}%
$^,$\footnote
{The centering step is vestigial: we can
  include it, ${\bf C} = {\bf S}^{-1/2}{\bf H(W-S)HS}^{-1/2}$, but it
  has no effect.  The discarded eigenvector (which corresponds to
  eigenvalue 0 and to a constant coordinate in the embedding)
  similarly is vestigial: we can discard it or not, and it doesn't
  affect pairwise distances among embedded points.  Interestingly,
  previous papers on the connection between Laplacian eigenmaps and
  the other algorithms mentioned here seem to contain an imprecision:
  they typically connect dropping the top (trivial) eigenvector with
  the centering step---e.g., see~\cite{lawrence-aistats11}. }%
$^,$\footnote{A minor difference is that LE does not scale
  its embedding using the eigenvalues of $\bf G$; the related
  \emph{diffusion maps} algorithm~\cite{Nadler05} does.} 
  %

 The weighted Gram matrix $\bf C$ can be related to a random walk on the graph defined by $\bf W$: if
we add the identity and make a similarity transform (both operations
that preserve the eigensystem), we get ${\bf S}^{-1/2}{\bf
  (C+I)S}^{1/2} = {\bf S}^{-1}{\bf W}$, a stochastic transition matrix
for the walk. 
  So, Laplacian eigenmaps can be viewed as trying to keep
points close together when they are connected by many short paths in
our graph.%
%
   %
  


\section{Bias and Instrumental Variables}
\label{sec:crosscov}
Kernel eigenmap methods are very good at dimensionality reduction when
the original data points sample a high-dimensional manifold relatively
densely, and when the noise in each sample is small compared to the
local curvature of the manifold (or when we don't care about
recovering curvature on a scale smaller than the noise).  
In practice, however, observations are frequently noisy.
Depending on the nature of the noise, manifold-learning 
algorithms applied to these datasets produce \emph{biased}
embeddings. See Figures 1--2, the ``noisy swiss rolls,'' for an example of how noise can bias manifold learning algorithms.
  
To see why, we examine PCA, a special case of manifold learning
methods, and look at why it produces biased embeddings in the presence
of noise. We first show how overcome this problem in the linear case,
and then use these same ideas to fix kernel PCA, a nonlinear
algorithm.
Finally, in Sec.~\ref{sec:2m}, we extend these ideas to fully general
kernel eigenmap methods.

\subsection{Bias in Finite-Dimensional Linear Models}
\label{sec:linear}
Suppose that
${\bf x}_i$ is a noisy view of some underlying
low-dimensional latent variable ${\bf z}_i$: ${\bf
  x}_i = {\bf M}{\bf z}_i+\epsilon_i$ for a linear transformation $\bf
M$ and 
i.i.d.\ zero-mean noise term $\epsilon_i$.  Without loss of generality, we assume
that ${\bf x}_i$ and ${\bf z}_i$ are centered, and that
$\text{Cov}[{\bf z}_i]$ and $\bf M$ both have full column rank: 
 any component of ${\bf z}_i$ in the nullspace of \mbox{$\bf M$ doesn't
 affect ${\bf x}_i$.}

In this case, PCA  
on ${\bf X}$  will
generally \emph{fail} to recover ${\bf Z}$: the
expectation of ${\bf \hat \Sigma}_{XX} = \frac{1}{n}{\bf X}{\bf X}\tps$ is 
${\bf M} \,\text{Cov}[{\bf z}_i]\,{\bf M}\tps +\text{Cov}[\epsilon_i]$, 
while we need ${\bf M} \,\text{Cov}[{\bf z}_i]\,{\bf M}\tps$
to be able to recover a
transformation of $\bf M$ or $\bf Z$. 
The unwanted term $\text{Cov}[\epsilon_i]$
will, in general, affect all eigenvalues and eigenvectors of
${\bf \hat \Sigma}_{XX}$, causing us to recover a \emph{biased} answer even
in the limit of infinite data. 



\subsubsection{Instrumental Variables}
\label{sec:instrumental}
We can fix this problem for linear embeddings: instead of  plain
PCA, we can use what might be called \emph{two-subspace} PCA\@.  This method
finds a statistically consistent solution through the use
of an \emph{instrumental variable}~\cite{Pearl2000,Jolliffe2002}, an
observation ${\bf y}_i$ that is correlated with the true latent
variables, but uncorrelated with the noise in ${\bf x}_i$.  

Importantly, picking an instrumental variable is
\emph{not} merely a statistical aid, but rather a \emph{value
  judgement} about the nature of the latent variable and the noise in
the observations.  In particular, we are \emph{defining} the noise to
be that part of the variability which is uncorrelated with the
instrumental variable, and the signal to be that part which is
correlated.


In our example above, a good instrumental variable ${\bf y}_i$ is a
different (noisy) view of the same underlying low-dimensional latent
variable: ${\bf y}_i = {\bf N}{\bf z}_i+\zeta_i$ for some
full-column-rank linear transformation $\bf N$ and i.i.d.\ zero-mean
noise term $\zeta_i$.
The expectation of the empirical cross covariance ${\bf \hat
  \Sigma}_{XY} = \frac{1}{n}\bf X Y\tps$ is then ${\bf
  M}\,\text{Cov}({\bf z}_i)\,{\bf N}\tps$: the noise terms, being
independent and zero-mean, cancel out.  (And the variance of each
element of ${\bf \hat \Sigma}_{XY}$ goes to 0 as $n\to\infty$.) 

In this case, we can identify the embedding
by computing the singular value decomposition (SVD) of the covariance:
 let 
${\bf UDV\tps} = {\bf \hat \Sigma}_{XY}$, where $\bf U$ and $\bf V$ are
orthonormal and $\bf D$ is diagonal.  
To reduce noise, we can keep just
 the columns of $\bf
U$, $\bf D$, and $\bf V$ which correspond to the top $k$ largest
singular values
(diagonal entries of $\bf D$): $\langle {\bf U},{\bf
  D},{\bf V}\rangle = \text{SVD}({\bf \hat \Sigma}_{XY},k)$.
If we set $k$ to be the true dimension of $\bf z$, then as $n\to\infty$, $\bf U$ will converge to an
orthonormal basis
for the range of $\bf M$, and $\bf V$ will converge to an orthonormal
basis for the range of $\bf N$.  The corresponding embeddings are then
given by $\bf U\tps X$ and $\bf V\tps Y$.

Interestingly, we can equally well view ${\bf x}_i$ as an instrumental
variable for ${\bf y}_i$: we simultaneously find consistent
embeddings of both ${\bf x}_i$ and ${\bf y}_i$, using each to unbias
the other.


\subsubsection{Whitening: Reduced-Rank Regression and Cannonical Correlation Analysis }\label{sec:whitening}
Going beyond two-subspace PCA, there are a number of interesting
spectral decompositions of cross-covariance matrices that involve
transforming the variables ${\bf x}_i$ and ${\bf y}_i$ before applying
a singular value decomposition~\cite{Jolliffe2002}. For example, in
reduced-rank regression~\cite{reinsel98,Jolliffe2002}, we want to
estimate $\mathbb{E}\left [{\bf x}_i\mid {\bf y}_i\right ] $.  Define
${\bf \hat \Sigma}_{YY} = \frac{1}{n} {\bf Y Y\tps}$ and ${\bf \hat
  \Sigma}_{XY} = \frac{1}{n} {\bf X Y\tps}$.  Then the ordinary
regression of $\bf Y$ on $\bf X$ is ${\bf \hat \Sigma}_{XY}( {\bf \hat
  \Sigma}_{YY} + \eta {\bf I})^{-1}{\bf Y}$, where the regularization
term $\eta {\bf I} \,\, (\eta > 0)$ ensures that the matrix inverse is
well-defined.  For a reduced-rank regression, we instead project onto
the set of rank-$k$ matrices: $\langle {\bf U},{\bf
  D},{\bf V}\rangle = \text{SVD}({\bf \hat \Sigma}_{XY} ( {\bf \hat \Sigma}_{YY} + \eta {\bf
  I})^{-1}{\bf Y},k)$.  In our
example above, so long as we let $\eta \rightarrow 0$ as $n\to\infty$,
$\bf U$ will again converge to an orthonormal basis for the range of
$\bf M$.

To connect back to two-subspace PCA,
we can show that RRR is the same as two-subspace PCA if we first
\emph{whiten} $\bf Y$.  That is,
we can equivalently define $\bf U$ for RRR to be the top $k$ left singular
vectors of the covariance between $\bf X$ and the {whitened}
instruments ${\bf Y}_w = ( {\bf \hat \Sigma}_{YY} +
\eta {\bf I}
)^{-1/2}\bf Y$.
(Here, ${\bf A}^{1/2}$ stands for the symmetric square root of $\bf
A$, which is guaranteed to exist and be invertible if $\bf A$ is
symmetric and positive definite.)
 Whitening means transforming a covariance matrix toward
the identity: if $\eta=0$ and ${\bf \hat\Sigma}_{YY}$ has full rank, then
$\mathbb E[{\bf Y}_w{\bf Y}_w\tps] =  {\bf \hat \Sigma}_{YY}^{-1/2}
{\bf \hat \Sigma}_{YY}  {\bf \hat \Sigma}_{YY}^{-1/2}={\bf I}$, while
if $\eta$ is near 0, then $\mathbb E[{\bf Y}_w{\bf
  Y}_w\tps]\approx{\bf I}$.

For symmetry, we can whiten both $\bf X$ and $\bf Y$ before computing
their covariance.  An SVD of the resulting doubly-whitened
cross-covariance matrix
$({\bf \hat \Sigma}_{XX} + \eta {\bf I})^{-1/2}
{\bf \hat \Sigma}_{XY}({\bf \hat \Sigma}_{YY} + \eta {\bf I})^{-1/2}$
is called canonical correlation
analysis~\cite{Hotelling1935}, and the resulting singular values are
called canonical correlations.

\subsection{Bias in Learning Nonlinear Models}
We now extend the analysis of Section~\ref{sec:linear} to
\emph{nonlinear} models.  We assume noisy observations ${\bf x}_i =
f({\bf z}_i)+\epsilon_i$, where ${\bf z}_i$ is the desired
low-dimensional latent variable, $\epsilon_i$ is an i.i.d.\ 
noise term, and $f$ is a smooth function with smooth inverse (so that
$f({\bf z}_i)$ lies on a manifold).  Our goal is to recover $f$ and
${\bf z}_i$ up to identifiability.

Kernel PCA (Sec.~\ref{sec:KPCA}) is a common approach to this problem.
In the realizable case, kernel PCA gets the right answer: that is,
suppose that ${\bf z}_i$ has dimension $k$, and that we have at least $k$
independent samples.  And, suppose that $\phi(f({\bf z}))$ is a
linear function of $\bf z$.  Then,
the Gram matrix or the covariance ``matrix'' will have rank $k$, and
we can reconstruct a basis for the range of $\phi\circ f$ from the top $k$
eigenvectors of the Gram matrix.  (Similarly, if $\phi\circ f$ is near
linear and the variance of $\epsilon_i$ is small, we can expect kernel
PCA to work well, if not perfectly.)

However, just as PCA recovers a biased answer in the finite
dimensional case when the variance of $\epsilon_i$ is nonzero, kernel
PCA will also recover a \emph{biased} answer in this case, even in the
limit of infinite data. The bias of kernel PCA follows immediately
from the example at the beginning of Section~\ref{sec:crosscov}: if we
use a linear kernel, kernel PCA will simply reproduce the bias of
ordinary PCA\@.

\subsubsection{Instrumental Variables}
By analogy to two-subspace PCA, a 
a natural generalization 
of kernel PCA is \emph{two-subspace kernel PCA}, which we can accomplish via
a kernelized SVD of a \emph{cross-covariance} operator in
Hilbert space.  Given a joint distribution $\PP[X,Y]$ over two
variables $X$ on $\Xcal$ and $Y$ on $\Ycal$, with feature maps $\phi$
and $\upsilon$ (corresponding to kernels $K_{\bf x}$ and
$K_{\bf y}$), the cross-covariance operator ${\bf
  \Sigma}_{XY}$ is $\mathbb E[\phi({\bf
  x})\otimes\upsilon({\bf y}) ]$.  
  %
  %
%
The cross-covariance operator reduces to an
ordinary cross-covariance matrix in the finite-dimensional case; in
the infinite-dimensional case, it can be viewed as a \emph{kernel mean
  map} descriptor~\cite{SmoGreSonSch07b} for the joint distribution
$\PP[X,Y]$.

The concept of a cross-covariance operator is helpful because it allows us to extend the methods of instrumental variables to
infinite dimensional RKHSs.
In our example above, a good instrumental variable ${\bf y}_i$ is a
different (noisy) view of the same underlying low-dimensional latent
variable: ${\bf y}_i = g({\bf z}_i)+\zeta_i$ for some
smoothly invertible function $g$ and i.i.d.\ zero-mean
noise term $\zeta_i$.

We proceed now to derive the kernel SVD for a cross-covariance operator.%
\footnote{The kernel SVD algorithm previously appeared as an intermediate
  step in~\cite{Song:2010fk,FukBacGre05}; here we generalize the
  algorithm to \emph{weighted} cross covariance operators in Hilbert
  space and give a more complete description, both because the method
  is interesting in its own right, and because this generalized SVD
  will serve as a step in our two-manifold algorithms.}  We show below
(Sec.~\ref{sec:KSVD}) that the kernel SVD of the cross-covariance
operator leads to a consistent estimate of the shared latent variables
in two-manifold problems which satisfy appropriate assumptions.


Conceptually, our inputs are ``matrices'' $\bf\Phi$ and $\bf\Upsilon$ whose columns
are respectively $\phi({\bf x}_i)$ and $\upsilon({\bf y}_i)$, along with
data weights ${\bf P}_X$ and ${\bf P}_Y$.  The centered empirical
covariance operator is then ${\bf \hat
  \Sigma}_{XY}=\frac{1}{n}{\bf ( \Phi H) ( \Upsilon H)\tps}$.  The
goal of the kernel SVD is then to factor $\hat{\bf  \Sigma}_{XY}$ (or
possibly the weighted variant ${\hat{\bf  \Sigma}_{XY}^{\bf P}} =
\frac{1}{n}({\bf  \Phi H}) {\bf P}_X{\bf P}_Y{\bf ( \Upsilon H})\tps$)
so that we can recover the desired bases for $\phi({\bf x}_i)$ and
$\upsilon({\bf y}_i)$. 

However, this conceptual algorithm
is impractical, since ${\bf \hat \Sigma}_{XY}$ can be high- or
infinite-dimensional.
%
Instead, we can perform an SVD on the covariance operator in Hilbert
space via a trick analogous to kernel PCA\@.  We first show how to
perform an SVD on a covariance matrix using Gram matrices in
finite-dimensional space, and then we extend the method to infinite
dimensional spaces in Section~\ref{sec:KSVD}\@.

\subsubsection{SVD via Gram matrices}
\label{sec:euclidSVD}
We start by looking at a Gram matrix formulation of finite dimensional SVD\@.
In standard SVD, the singular values of ${\bf \hat \Sigma}_{XY} = \frac{1}{n} \bf (X H) (Y H)\tps$ are the
square roots of the eigenvalues of ${\bf \hat \Sigma}_{XY}{\bf \hat
  \Sigma}_{YX}$ (where ${\bf \hat
  \Sigma}_{YX}={\bf \hat
  \Sigma}_{XY}\tps$), and
the left singular vectors are defined to be the corresponding
eigenvectors.  
We can find identical eigenvectors and eigenvalues through centered
Gram matrices ${\bf B}_X = \frac{1}{n}\bf (XH)\tps(XH)$ and ${\bf B}_Y = \frac{1}{n}\bf (YH)\tps(YH)$.
Let ${\bf v}_i$ be the right eigenvector of ${\bf B}_Y{\bf B}_X$ so that ${\bf B}_{Y}{\bf B}_X {\bf v}_i = \lambda_i {\bf v}_i$.
Premultiplying by $\bf (XH)$ yields
\begin{align*}
\frac{1}{n^2}{\bf (XH)  (YH)\tps (YH) (XH)\tps(XH) v}_i=\lambda_i {\bf (XH) v}_i
\end{align*}
and regrouping terms gives us ${\bf\hat \Sigma}_{XY}{\bf \hat \Sigma}_{YX}{\bf w}_i = \lambda_i{\bf w}_i$
where ${\bf w}_i = {\bf (XH)} {\bf v}_i$. So, $\lambda_i$ is an eigenvalue of ${\bf\hat \Sigma}_{XY}{\bf \hat \Sigma}_{YX}$, $\sqrt{\lambda_i}$ is a singular value of ${\bf \hat \Sigma}_{XY} $, and
$
{\bf (XH)} {\bf v}_i\lambda_i^{-1/2}
$
is the corresponding {unit length} left singular vector.
An analogous argument shows
that, if ${\bf w}_i'$ is a unit-length right singular vector of ${\bf \hat \Sigma}_{XY} $, then
${\bf w}_i'={\bf (YH)}{\bf v}_i'\lambda_i^{-1/2}$, where ${\bf v}_i'$
is a unit-length left
eigenvector of ${\bf B}_Y{\bf B}_X$.
%
%
Furthermore, we can easily incorporate weights into SVD by using weighted matrices ${\bf C}_X = {\bf P}_X{\bf B}_X{\bf P}_X$  and ${\bf C}_Y = {\bf P}_Y{\bf B}_Y{\bf P}_Y$ in place of ${\bf B}_X$ and  ${\bf B}_Y$.

\subsubsection{Two-subspace PCA in RKHSs}
\label{sec:KSVD}
The machinery developed in Section~\ref{sec:euclidSVD} allows us to solve the two-subspace kernel PCA problem by computing the singular values of the weighted empirical covariance operator ${\bf \hat \Sigma^P}_{XY}$. 
We define ${\bf G}_X$ and ${\bf G}_Y$ to be the Gram
matrices whose elements are $K_{\bf x}({\bf x}_i,{\bf x}_j) $ and
$K_{\bf y}({\bf y}_i,{\bf y}_j)$ respectively, 
and then compute the
eigendecomposition of ${\bf C}_{Y}{\bf C}_{X} = {\bf (P}_Y{\bf HG}_Y{\bf HP}_Y)({\bf P}_X{\bf H G}_X{\bf HP}_X)$.  This method avoids any
computations in infinite-dimensional spaces; and, it gives us compact
representations of the left and right singular vectors.  E.g., if ${\bf
  v}_i$ is a right eigenvector of $ {\bf C}_{Y}{\bf C}_{X} $, then the corresponding singular vector is ${\bf w}_i =
\sum_j {\bf v}_{i,j}\,\upsilon({\bf x}_j - {\bf \bar x})p^x_j$, where $p_j^x$ is the weight assigned to data point ${\bf x}_j$ .


Under appropriate assumptions, we can show that the SVD of the
empirical cross-covariance operator ${\bf \hat \Sigma}_{XY} =
\frac{1}{n}{\bf \Phi}{\bf H \Upsilon}\tps$ converges to the desired
value.  Suppose that $\mathbb E[\phi({\bf x}_i)\mid {\bf z}_i]$ is a
linear function of ${\bf z}_i$, and similarly, that $\mathbb
E[\upsilon({\bf y}_i)\mid {\bf z}_i]$ is a linear function of ${\bf
  z}_i$.%
\footnote{The assumption of linearity is restrictive, but appears
  necessary: in order to learn a representation of a manifold using
  factorization-based methods, we need to pick a kernel which flattens
  out the manifold into a subspace.  This is why kernel eigenmap
  methods are generally more successful than plain kernel PCA:
  by learning an appropriate kernel, they are able to adapt their
  nonlinearity to the shape of the target manifold.}  
The noise terms
$\phi({\bf x}_i)-\mathbb E[\phi({\bf x}_i)\mid {\bf z}_i]$ and
$\upsilon({\bf y}_i)-\mathbb E[\upsilon({\bf y}_i)\mid {\bf z}_i]$ are
by definition zero-mean; and they are independent of each other, since
the first
depends only on $\epsilon_i$ and the second only on $\zeta_i$.  So,
the noise terms cancel out, and the expectation of ${\bf \hat \Sigma}_{XY}$ is the true covariance
${\bf \Sigma}_{XY}$.  If we additionally assume that the noise terms
have finite variance, the product-RKHS norm of the error ${\bf \hat
  \Sigma}_{XY} - {\bf \Sigma}_{XY}$ vanishes as $n\to\infty$.

The remainder of the proof follows from the proof of Theorem 1
in~\cite{Song:2010fk} (the convergence of the empirical estimator of
the kernel covariance operator).  In particular, the top $k$ left
singular vectors of ${\bf \hat \Sigma}_{XY}$ converge to a basis for
the range of $\mathbb E[\phi({\bf x}_i)\mid {\bf z}_i]$ (considered as
a function of ${\bf z}_i$); similarly, the top right singular vectors
of ${\bf \hat \Sigma}_{XY}$ converge to a basis for the range of
$\mathbb E[\upsilon({\bf y}_i)\mid {\bf z}_i]$.

\subsubsection{Whitening and Kernel SVD}
\label{sec:kwhitening}
Just as one can use kernel SVD to solve the two-subspace kernel PCA problem for high-
or infinite-dimensional feature space, one can also compute the kernel
versions of CCA and RRR\@. (Kernel CCA 
is a well-known algorithm, though our formulation here is different
than in~\cite{FukBacGre05}, and kernel RRR is to our knowledge novel.) Again, these problems can be solved by pre-whitening the feature space before applying kernel SVD.

%
%
%
%
%

To compute a RRR from centered covariates $\bf H \Upsilon$
in Hilbert space to centered responses $\bf H \Phi$ in Hilbert space,
we find the kernel SVD of the $\bf H \Upsilon$-whitened covariance
matrix: first we define ${\bf B}_X = {\bf HG}_{X}{\bf H}$ and ${\bf B}_Y = {\bf
HG}_Y{\bf H}$; next we compute ${\bf B}_X{\bf B}_Y ({\bf B}_Y^2 + \eta {\bf
  I})^{-1} {\bf B}_Y $; and, finally, we perform kernel SVD by finding the singular
  value decomposition of ${\bf B}_X{\bf B}_Y ({\bf B}_Y^2 + \eta {\bf
  I})^{-1} {\bf B}_Y $.

  Similarly, for CCA in Hilbert space, we find the
kernel SVD of the $\bf H \Upsilon$- and $\bf H \Phi$-whitened
covariance matrix: ${\bf  B}_X({\bf  B}_X^2 + \eta {\bf I})^{-1}{\bf B}_X{\bf B}_Y ({\bf B}_Y^2 + \eta {\bf I})^{-1}{\bf B}_Y $.
  
If data weights are present, we use instead weighted centered Gram matrices ${\bf C}_{X} = {\bf P}_X{\bf  B}_X{\bf P}_X$ and ${\bf C}_{Y} = {\bf P}_Y {\bf B}_Y{\bf P}_Y$. Then, to generalize RRR and CCA to RKHSs, we can compute the eigendecompositions of 
Equations~\ref{eq:rrr}--b, respectively:
\begin{subequations}
\begin{align}
\label{eq:rrr}
{\bf C}_X{\bf C}_Y ({\bf C}_Y^2 + \eta {\bf I})^{-1}{\bf C}_Y\qquad\qquad\quad\,\,\,\\ \label{eq:cca}
{\bf C}_X({\bf C}_X^2 + \eta {\bf I})^{-1}{\bf C}_X{\bf C}_Y ({\bf C}_Y^2 + \eta {\bf I})^{-1} {\bf C}_Y
\end{align}
\end{subequations}
The consistency of both approaches follows directly from the
consistency of kernel SVD, so long as we let $\eta \to 0$ as $n\to\infty$~\cite{song09a}.

%
%

\section{Two-Manifold Problems}
\label{sec:2m}
Now that we have extended the instrumental variable idea
to RKHSs, we can also expand the scope of manifold learning to \emph{two-manifold problems}, where
we want to simultaneously learn two manifolds for two covarying lists
of observations, each corrupted by uncorrelated noise.%
\footnote{The uncorrelated noise assumption is extremely mild:
  if some latent variable causes correlated changes in our
  measurements on the two manifolds, then we are making the definition
  that it is part of the desired signal to be recovered.  No other
  definition seems reasonable: if there is no
  difference in statistical behavior between signal and noise, then it
  is impossible to use a statistical method to separate signal from
  noise.}
The idea is simple: we view manifold learners as constructing Gram
matrices, then apply the RKHS instrumental variable idea of
Sec.~\ref{sec:crosscov}.  As we will see below, this procedure allows us
to regain good performance when observations are noisy.



\begin{figure}[t]\begin{center}
    \includegraphics[width=1\columnwidth]{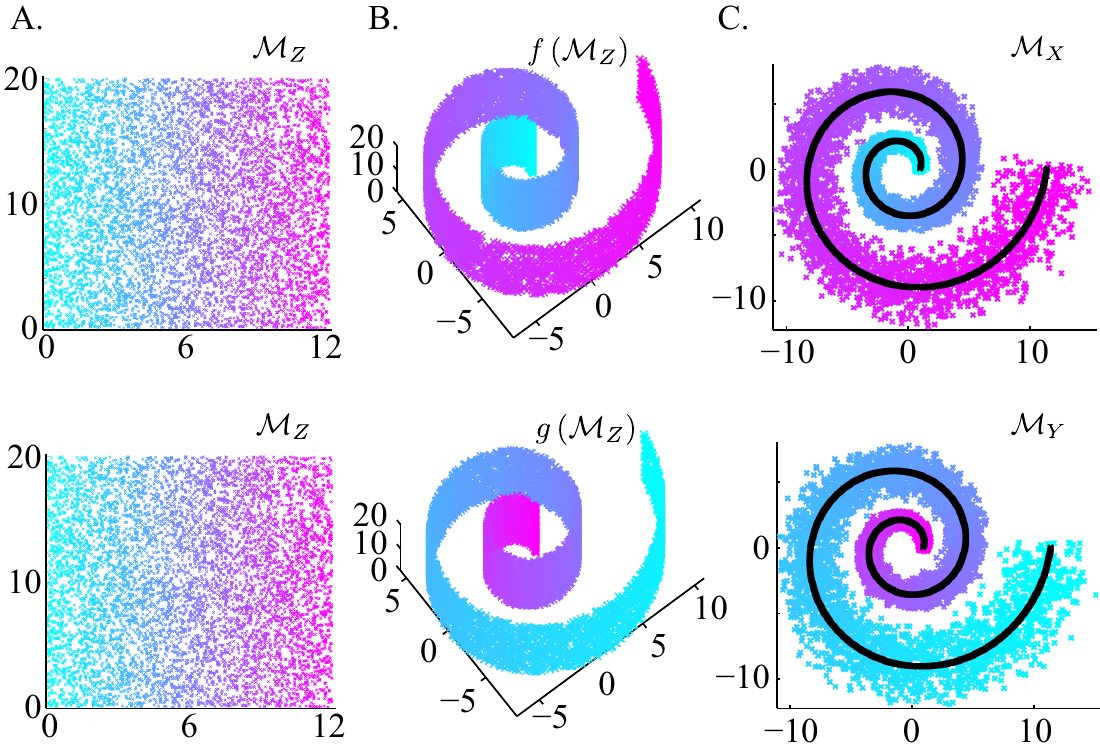}
    \caption{The Noisy Swiss Rolls.  We are given two sets of
      3-d observations residing on two different
       manifolds $\mathcal{M}_X$ and $\mathcal{M}_Y$.
      The latent \emph{signal} $\mathcal{M}_Z$ is 2-d, but
      $\mathcal{M}_X$ and $\mathcal{M}_Y$ are each corrupted by
      3-d \emph{noise}. (A) 5000 data points sampled from
       $\mathcal{M}_Z$. (B) The
      functions $f\left( \mathcal{M}_Z\right)$ and 
      $g\left(\mathcal{M}_Z\right)$  ``roll'' the manifold in
      3-dimensional space, two different ways, generating two
      different manifolds in observation space. (C) Each set of
      observations is then perturbed by 3-d noise
      (showing 2 dimensions only), resulting in 3-d
      manifolds $\mathcal{M}_X$ and $\mathcal{M}_Y$. The black lines
      indicate the location of the submanifolds from (B). 
    }
    \label{fig:manifoldExample}
    \end{center}
\end{figure}

Suppose we are given two set of observations residing on (or
near) two different manifolds: ${\bf x}_1, \hdots, {\bf x}_n \in \mathbb{R}^{d_1}$ on
a manifold $\mathcal{M}_X$ and ${\bf y}_1, \hdots, {\bf y}_n \in \mathbb{R}^{d_2}$ on a manifold $\mathcal{M}_Y$. Further suppose that both ${\bf x}_i$ and ${\bf y}_i$ are \emph{noisy}
functions of a latent variable ${\bf z}_i$, itself residing on a
latent $k$-dimensional manifold $\mathcal{M}_Z$: ${\bf x}_i = f({\bf z}_i)+\epsilon_i$
and ${\bf y}_i = g({\bf z}_i)+\zeta_i$.  We assume that the
functions $f$ and $g$ are smooth, so that $f({\bf z}_i)$ and $g({\bf
  z}_i)$ trace out submanifolds $f(\mathcal{M}_Z) \subseteq \mathcal{M}_X$
and $g(\mathcal{M}_Z) \subseteq \mathcal{M}_Y$.  We further assume that the
noise terms $\epsilon_i$ and $\zeta_i$ move ${\bf x}_i$ and ${\bf
  y}_i$ \emph{within} their respective manifolds $\mathcal{M}_X$ and
$\mathcal{M}_Y$: this assumption is without loss of generality, since we
can can always increase the dimension of the manifolds $\mathcal{M}_X$
and $\mathcal{M}_Y$ to allow an arbitrary noise term. See Figure~\ref{fig:manifoldExample}
for an example.


If the variance of the noise terms $\epsilon_i$ and $\zeta_i$ is
too high, or if $\mathcal{M}_X$ and $\mathcal{M}_Y$ are higher-dimensional
than the latent $\mathcal{M}_Z$ manifold (i.e., if the noise terms move
${\bf x}_i$ and ${\bf y}_i$ away from $f(\mathcal{M}_Z)$ and $g(\mathcal{M}_Z)$),
 then it may be difficult to reconstruct $f(\mathcal{M}_Z)$ or
$g(\mathcal{M}_Z)$ separately from ${\bf x}_i$ or ${\bf y}_i$.  
Our goal,
therefore, is to use ${\bf x}_i$ and ${\bf y}_i$ together to
reconstruct both manifolds simultaneously: the extra
information from the correspondence between ${\bf x}_i$ and ${\bf
  y}_i$ will make up for noise, allowing
success in the two-manifold problem where the individual one-manifold
problems are intractable.

Given samples of $n$ i.i.d. pairs $\{ {\bf x}_i, {\bf y}_i\}_{i=1}^n$
from two manifolds, we propose a two-step spectral learning algorithm
for two-manifold problems: first, use either a given kernel or an
ordinary 
one-manifold algorithm such as LE or LLE to compute weighted centered
Gram matrices ${\bf C}_X $ and ${\bf C}_Y$ from ${\bf x}_i$ and ${\bf
  y}_i$ separately. Second, use
 use one of the cross-covariance methods from
Section~\ref{sec:kwhitening}, such as kernel SVD, to recover the embedding
of points in $\mathcal{M}_Z$. The procedure, called instrumental eigenmaps, is summarized in Algorithm~\ref{alg:2manifold}.

\begin{algorithm}[t] 
\caption{Instrumental Eigenmaps}
\textbf{In}: $n$ \emph{i.i.d.} pairs of observations $\{ {\bf x}_i,
{\bf y}_i\}_{i=1}^n$\\ 
\textbf{Out}:  embeddings ${\bf E}_X$ and ${\bf E}_Y$ \vspace{-3mm}\\
  \begin{algorithmic}[1]  \label{alg:2manifold}
  \STATE Compute centered Gram matrices: ${\bf B}_X$ and ${\bf B}_Y$ and data weights ${\bf P}_X$  and ${\bf P}_Y$ from ${\bf x}_{1:n}$ and  ${\bf y}_{1:n}$
 \STATE Compute \emph{weighted}, centered Gram matrices:\\  ${\bf C}_X = {\bf P}_X {\bf B}_X {\bf P}_X$  and ${\bf C}_Y = {\bf P}_Y {\bf B}_Y {\bf P}_Y$
 \STATE Perform a singular value decomposition and truncate the top $k$ singular values:\\  $\langle {\bf U, \Lambda, V\tps} \rangle = \mathop{\text{SVD}}\!\left ( {\bf
    C}_X{\bf C}_Y, k \right )$
 \STATE Find the embeddings from the singular values:\\
 ${\bf E}_{X} = {\bf P}_X^{1/2}{\bf U}_{2:k+1}{\bf \Lambda}_{2:k+1}^{1/2}$ and\\ ${\bf E}_Y = {\bf P}_Y^{1/2}{\bf V}_{2:k+1}{\bf \Lambda}_{2:k+1}^{1/2}$
  \end{algorithmic}
  \end{algorithm}

For example, if we are using LE (Section~\ref{sec:Manifold}), then
${\bf C}_X = {\bf S}^{-1/2}_X ({\bf W}_X-{\bf S}_X) {\bf S}^{-1/2}_X$ and
${\bf C}_Y = {\bf S}^{-1/2}_Y ({\bf W}_Y-{\bf S}_Y) {\bf S}^{-1/2}_Y$,
where ${\bf W}_X$ and ${\bf S}_X$ are computed from ${\bf x}_i$,
and ${\bf W}_Y$ and ${\bf S}_Y$ are computed from ${\bf y}_i$. We
then perform a singular value decomposition and truncate to the top
$k$ singular values:
\begin{align*}
\langle {\bf U, \Lambda, V\tps} \rangle = \mathop{\text{SVD}}\!\left ( {\bf
    C}_X{\bf C}_Y, k \right )
\end{align*}
The embeddings for ${\bf x}_i$ and ${\bf y}_i$ are then given by 
\begin{subequations}
\begin{align*}
&{\bf S}_X^{1/2}{\bf U}_{2:k+1}{\bf \Lambda}_{2:k+1}^{1/2} \quad \text{and} \quad {\bf S}_Y^{1/2}{\bf V}_{2:k+1}{\bf \Lambda}_{2:k+1}^{1/2}.
\end{align*}
\end{subequations}

\begin{figure}[t]\begin{center}
    \includegraphics[width=.7\columnwidth]{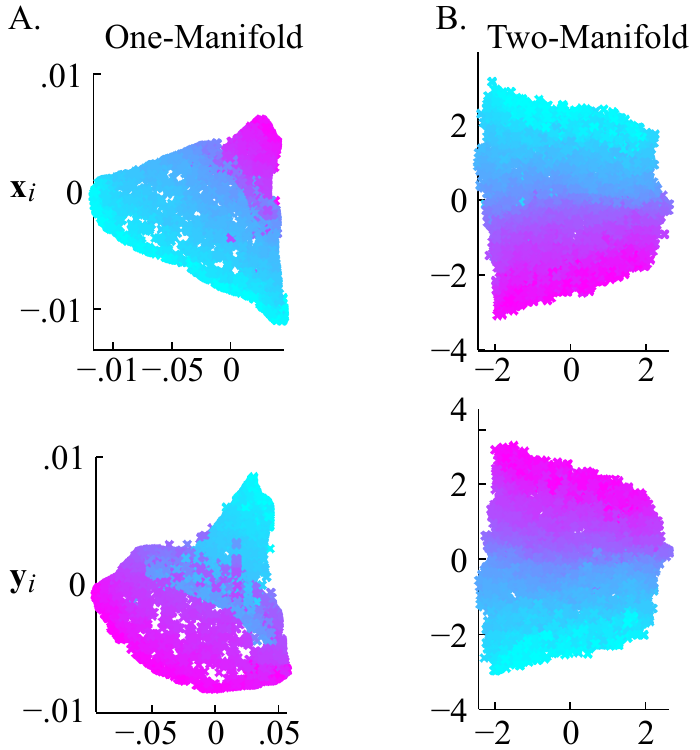}
    \caption{Solving the Noisy Swiss Roll two-manifold problem (see Fig.~\ref{fig:manifoldExample} for setup).  Top graphs show   embeddings of ${\bf x}_i$, bottom graphs show embeddings of ${\bf y}_i$. 
    (A) 2-dimensional embeddings found by LE. The best results were obtained by setting the number of nearest neighbors to 5. Due to the large amounts of noise, the separately learned embeddings do not accurately reflect the latent 2-dimensional manifold. 
    (B) The embeddings learned from the left and right eigenvectors of ${\bf C}_X{\bf C}_Y$ closely match the original data sampled from the true manifold.  By treating the learning problem as a two-manifold problem, noise disappears in expectation and the latent manifold is recovered accurately.}
    \label{fig:manifoldExample2}
    \end{center}
\end{figure}

Computing eigenvalues of ${\bf C}_X {\bf C}_Y$ 
 instead of ${\bf C}_X$ or ${\bf C}_Y$ alone will alter
the eigensystem: it will promote directions within each individual
learned manifold that are useful for predicting coordinates on the
other learned manifold, and demote directions that are not useful.  As shown
in Figure~\ref{fig:manifoldExample2}, this effect strengthens our ability 
to recover relevant dimensions in the face of noise.

%

%

\section{Two-Manifold Detailed Example: Learning Dynamical Systems}
\label{sec:Subspace}
A longstanding goal in machine learning and robotics has been to learn
accurate, economical models of dynamical systems directly from
observations. This task requires two related subtasks: 1) learning a
low dimensional state space, which is often known to lie on a
\emph{manifold}; and 2) learning the system dynamics. 
 We propose tackling this problem by combining
spectral learning algorithms for non-linear dynamical
systems~\cite{zhang09,Siddiqi10a,Boots2010b,Boots2011a,Boots-online-psr,Song:2010fk}
with two-manifold methods.  

Any of the above-cited spectral learning algorithms can be combined
with two-manifold methods.  Here, we focus on one specific example: we
show how to combine HSE-HMMs~\cite{Song:2010fk}, a powerful
nonparametric approach to modeling dynamical systems, with manifold
learning.  Specifically, we look at one important step in the HSE-HMM
learning algorithm: the original approach uses kernel SVD to discover
a low-dimensional state space, and we propose replacing the kernel SVD with
a two-manifold method.
%
%
We demonstrate that the
resulting \emph{manifold} HSE-HMM can outperform standard HSE-HMMs
(and many other well-known methods for learning dynamical systems) on
a difficult real-world example: the manifold HSE-HMM accurately
discovers a curved low-dimensional manifold which contains the state
space, while other methods discover only a (potentially much higher-dimensional) subspace which contains
this manifold.

 \subsection{Hilbert Space Embeddings of HMMs}

\label{sec:stateSpace}
The key idea behind spectral learning of dynamical systems is that a
good latent state is one that lets us predict the future.  HSE-HMMs
implement this idea by finding a low-dimensional embedding of the
\emph{conditional probability distribution} of sequences of future
observations, and using the embedding coordinates as
state. Song et al.~\cite{Song:2010fk} suggest finding this low-dimensional state
space as a {subspace} of an infinite dimensional RKHS.

Intuitively, we might think that we could find the best state space by
performing PCA or kernel PCA of sequences of future observations.  That is,
we would sample $n$ sequences of future observations ${\bf x}_1,
\hdots, {\bf x}_n \in \mathbb{R}^{d_1}$ from a dynamical system.  (If
our training data consists of a single long sequence of observations,
we can collect our sample by picking $n$ random time steps
$t_1,\ldots,t_n$.  Each sample ${\bf x}_i$ is then a sequence of
observations starting at time $t_i$.)  We would then construct a Gram
matrix ${\bf G}_X$, whose $(i,j)$ element is $K_{\bf x}({\bf x}_i,
{\bf x}_j)$.  Finally, we would find the eigendecomposition of the centered Gram
matrix ${\bf B}_X = {\bf HG}_X{\bf H}$ as in Section~\ref{sec:KPCA}.  The
resulting embedding coordinates would be tuned to predict future
observations well, and so could be viewed as a good state space.


However, the state space found by kernel PCA is \emph{biased}: it typically
includes \emph{noise}, information that cannot be predicted from past
observations. We would like instead to find a low dimensional state
space that embeds the probability distribution of possible futures
\emph{conditioned} on the past. In particular, we want to find a state
space that is \emph{uncorrelated} with the noise in the future
observations.
From a two-manifold perspective, we view features of the past
as instrumental variables to unbias the future.

Therefore, in addition to sampling sequences of future observations,
we sample corresponding sequences of \emph{past} observations ${\bf
  y}_1, \hdots, {\bf y}_n \in \mathbb{R}^{d_2}$: sequence ${\bf y}_i$
ends at time $t_i-1$.  We then construct a Gram matrix ${\bf G}_Y$,
whose $(i,j)$ element is $K_{\bf y}({\bf y}_i, {\bf y}_j)$.  From ${\bf G}_Y$ we
construct the centered Gram matrix ${\bf B}_Y = {\bf HG}_Y{\bf H}$.
Finally, we identify the state space using a kernel SVD as in
Section~\ref{sec:KSVD}: $\langle {\bf U, \Lambda, V\tps} \rangle =
\text{SVD}( {\bf B}_X {\bf B}_Y,k)$ (there are no data weights for
HSE-HMMs).
%
%
The left singular ``vectors'' (reconstructed from $\bf U$ as in
Section~\ref{sec:KSVD}) now identify a subspace in which the system
evolves.  From this subspace, we can proceed to identify the
parameters of the dynamical system as in
Song et al.~\cite{Song:2010fk}.\footnote{Other approaches such as CCA and RRR
  have also been successfully used for finite-dimensional spectral
  learning algorithms~\cite{Boots2011a}, suggesting that kernel SVD
  could be replaced by kernel CCA or kernel RRR
  (Section~\ref{sec:kwhitening}).}

\subsection{Manifold HSE-HMMs}
\label{sec:manifoldHMM}
In contrast with HSE-HMMs, we are interested in modeling a dynamical
system whose state space lies on a low-dimensional manifold, even if
this manifold is curved to occupy a higher-dimensional subspace (an
example is given in Section~\ref{sec:slotcar}, below). 
We want to use this additional knowledge to constrain the learning
algorithm and produce a more accurate model for a given amount of
training data.

To do so, we replace the kernel SVD by a two-manifold method.  That
is, we learn weighted centered Gram matrices ${\bf C}_X$ and ${\bf
  C}_Y$ for the future and past observations, using a manifold method
like LE or LLE (see Section~\ref{sec:Manifold}).  Then we apply a
SVD to ${\bf C}_X{\bf C}_Y$ in order to
recover the latent state space.

\subsection{\mbox{Slotcar: A Real-World Dynamical System}}
\label{sec:slotcar}

\begin{figure*}[t]\begin{center}
    \includegraphics[width=2\columnwidth]{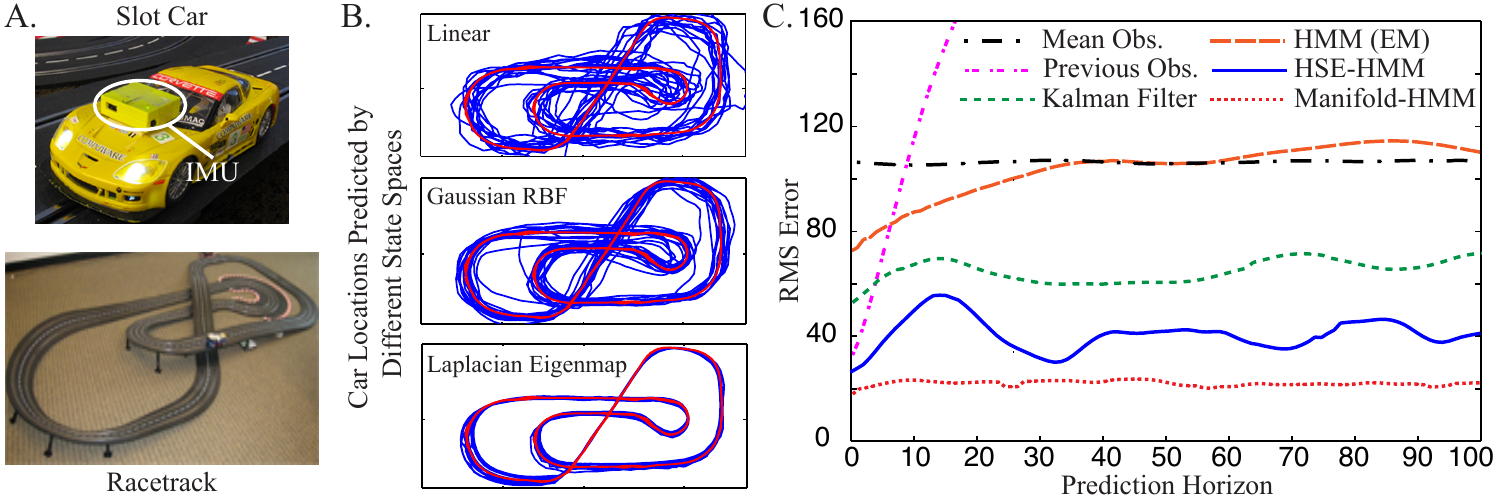}
    \end{center}
    \caption{Slot car with inertial measurement unit (IMU). (A) The slot car
      platform: the car and IMU (top) and racetrack (bottom). (B) A comparison of 
      training data embedded into the state space of three different learned models. 
      Red line indicates true 2-d position of the car over time, blue lines 
      indicate the prediction from state space. The top
      graph shows the Kalman filter state space (linear kernel), the middle 
      graph shows the HSE-HMM state space (Gaussian RBF kernel), and
      the bottom graph shows the manifold HSE-HMM state space (LE kernel).
      The LE kernel finds the best representation of the true manifold.
      (C) Root mean squared error for prediction (averaged over 250 trials) with different estimated models.  
    The HSE-HMM significantly outperforms
    the other learned models by taking advantage of the fact that the data we want to predict lies on a manifold. }
    \label{figure2}
\end{figure*}

Here we look at the problem of tracking and predicting the position of a slotcar with attached inertial measurement unit (IMU) racing around a track.
The setup consisted of a track and a miniature car (1:32 scale model)
guided by a slot cut into the track.  Figure~\ref{figure2}(A) shows setup.
At a rate of 10Hz, we extracted the
estimated 3-D acceleration and angular velocity of the car. An
overhead camera also supplied estimates of the 2-dimensional location
of the car. 

We collected 3000 successive observations while the slot car circled the track controlled by a constant policy (with varying speeds). The goal was to learn a dynamical model of the noisy IMU data, and, after filtering, to predict current and future 2-dimensional locations. 
We used the first 2000 data points as training data, and held out the
last 500 data points for testing the learned models. We trained four
models, and evaluated these models based on prediction accuracy, and,
where appropriate, the learned latent state.

First, we trained a 20-dimensional embedded HMM with the spectral
algorithm of Song et al.~\cite{Song:2010fk}, using sequences of 150 consecutive
observations. Following Song et al., we used Gaussian RBF kernels, 
setting the bandwidth parameter with the ``median trick'', and using
regularization parameter $\lambda=10^{-4}$. Second, we trained a similar
20-dimensional embedded HMM with LE kernels. The number of nearest
neighbors was selected to be 50, and the other parameters were set to
be identical to the first model.  (So, the only difference is that the
first model performs a kernel SVD to find the subspace on which the
dynamical system evolves, while the second model solves a two-manifold
problem.)  Third, for comparison's sake, we trained a 20-dimensional
Kalman filter using the N4SID algorithm~\cite{vanoverschee96book} with
Hankel matrices of 150 time steps; and finally, we learned a 20-state
discrete HMM (with 400 levels of discretization for observations) using
the EM algorithm.  

Before investigating the predictive accuracy of the learned dynamical
systems, we looked at the learned state space of the first three
models.  These models differ mainly in their kernel: Gaussian RBF,
learned manifold from Laplacian Eigenmaps, or linear.  As a test, we
tried to reconstruct the 2-dimensional locations of the car from each
of the three latent state spaces: the more accurate the learned state
space, the better we expect to be able to reconstruct the locations.
Results are shown in Figure~\ref{figure2}(B); the learned manifold is
the clear winner.


Finally we examined the prediction accuracy of each model. We performed filtering for different extents $t_1 = 100, \hdots , 350$, then
predicted the car location
for a further $t_2$ steps in
the future, for $t_2 = 1,\ldots, 100$. The root mean squared error of
this prediction in the 2-dimensional location space is plotted in Figure~\ref{figure2}(C). The Manifold HMM 
learned by the method detailed in Section~\ref{sec:manifoldHMM}
consistently yields lower prediction error for the duration of the
prediction horizon. This is important: not only does the manifold
HSE-HMM provide a model that better visualizes the data that we want to
predict, but it is significantly more accurate when filtering
and predicting than classic and state-of-the-art alternatives.

%

\section{Related Work}
\label{sec:relatedWork}
While preparing this manuscript, we learned of the simultaneous and independent work of Mahadevan et al.~\cite{Mahadevan2011}.  This paper defines one particular two-manifold algorithm, maximum covariance unfolding (MCU), by extending maximum variance unfolding; but, it does not discuss how to extend other one-manifold methods.  It also does not discuss any asymptotic properties of the MCU method, such as consistency.

A similar problem to the two-manifold problem is \emph{manifold alignment}~\cite{Ham05,Wang09}, which builds connections between two or more data sets by aligning their underlying manifolds. Generally, manifold alignment algorithms either first learn the manifolds separately and then attempt to align them based on their low-dimensional geometric properties, or they take the union of several manifolds and attempt to learn a latent space that preserves the geometry of all of them~\cite{Wang09}.  Our aim is different: we assume paired data, where manifold alignments do not; and, we focus on learning algorithms that \emph{simultaneously} discover manifold structure (as kernel eigenmap methods do) \emph{and} connections between manifolds (as provided by, e.g., a top-level learning problem defined between two manifolds).   

This type of interconnected learning problem has been explored before in a different context via  reduced-rank regression (RRR)~\cite{anderson51,izenman75,reinsel98} and sufficient dimension reduction (SDR)~\cite{li91,cook2002,fuk2004}.  RRR is a linear regression method that attempts to estimate a set of coefficients $\beta$ to predict response vectors ${\bf y}_i$ from covariate vectors ${\bf x}_i$ under the constraint that $\beta$ is low rank (and can therefore be factored). In SDR, the goal is to find a linear subspace of covariates ${\bf x}_i$ that makes response vectors ${\bf y}_i$ conditionally independent of the ${\bf x}_i$s. The formulation is in terms of conditional independence, and, unlike in RRR, no assumption is made on the form of the regression from ${\bf x}_i$ to ${\bf y}_i$. Unfortunately, SDR pays a price for the relaxation of the linear assumption: the solution to SDR problems usually requires  a difficult non-linear non-convex optimization. 

Manifold kernel dimension reduction~\cite{nilsson2007},
  finds an embedding of covariates ${\bf x}_i$ using a kernel eigenmap
  method, and then attempts to find a linear transformation of some of
  the dimensions of the embedded points to predict response variables
  ${\bf y}_i$. The response variables are constrained to be linear in
  the manifold, so the problem is quite different from a two-manifold
  problem.


There has also been some work on finding a mapping between manifolds~\cite{Steinke2010} and learning a dynamical system on a manifold~\cite{Tyagi2008}; however, in both of these cases it was assumed that the manifold was \emph{known}.

Finally, some authors have focused on the problem of
combining manifold learning algorithms with system identification
algorithms. For example, Lewandowski et al.~\cite{Lew2010}
introduces Laplacian eigenmaps that
accommodate time series data by picking neighbors based on temporal
ordering. Lin et al.~\cite{Lin06} and Li et al.~\cite{Li2007} propose
learning a piecewise linear model that approximates a non-linear
manifold and then attempt to learn the dynamics in the
low-dimensional space. Although these methods do a qualitatively
reasonable job of constructing a manifold, none of these papers
compares the predictive accuracy of its model to state-of-the-art
dynamical system identification algorithms.


\section{Conclusion}
In this paper we propose a class of problems called \emph{two-manifold problems},
where two sets of corresponding data points, generated from a single latent
manifold and corrupted by noise, lie on or near two different higher dimensional manifolds. We design algorithms
by relating two-manifold problems to cross-covariance operators in 
RKHSs, and show that these algorithms result in a significant
improvement over standard manifold learning approaches in the presence
of noise. This is an appealing result: manifold learning algorithms 
typically assume that observations are (close to) noiseless, an assumption
that is rarely satisfied in practice.

Furthermore, we demonstrate the utility of two-manifold problems by
extending a recent dynamical system identification algorithm to learn
a system with a state space that lies on a manifold. The resulting
algorithm learns a model that outperforms the current state-of-the-art
in terms of predictive accuracy. To our knowledge this is the first
combination of system identification and manifold learning that
accurately identifies a latent time series manifold {and} is
competitive with the best system identification algorithms at learning
accurate predictive models.

\section*{Acknowledgements}
Byron Boots and Geoffrey J. Gordon were supported by ONR MURI grant
number N00014-09-1-1052.  Byron Boots was supported by the NSF under
grant number EEEC-0540865.

\bibliography{references}  
\bibliographystyle{unsrt}
\end{document}